\documentclass[numbered]{trbunofficial}
\usepackage{graphicx}
\usepackage{amsmath}
\usepackage{booktabs}
\usepackage{tabularx}
\usepackage{array}
\usepackage{multirow}
\usepackage{threeparttable}
\usepackage{siunitx}
\usepackage{listings}
\usepackage{xcolor}
\usepackage{tikz}
\usetikzlibrary{arrows.meta,positioning,shapes.geometric,fit,decorations.pathreplacing}
\usepackage{microtype}
\usepackage{placeins}
\usepackage{fancyhdr}
\usepackage[
  authordate,
  backend=biber,
  sorting=nyt,
  sortcites=true,
  ibidtracker=false,
  giveninits=false,
  maxcitenames=2,
  mincitenames=1,
  uniquelist=false,
  maxbibnames=99,
  minbibnames=99,
  doi=only,
  url=true,
  eprint=false,
  isbn=false
]{biblatex-chicago}
\definecolor{PromptGray}{HTML}{F7F8FA}
\definecolor{Muted}{HTML}{5B6573}

\newcolumntype{Y}{>{\raggedright\arraybackslash}X}
\newcolumntype{C}{>{\centering\arraybackslash}X}

\newcommand{\mi}{\mathrm{mi}}
\newcommand{\minutes}{\mathrm{min}}

\lstdefinestyle{outputschema}{
  basicstyle=\ttfamily\small,
  frame=single,
  breaklines=true,
  columns=fullflexible,
  keepspaces=true,
  showstringspaces=false
}

\lstdefinestyle{promptblock}{
  basicstyle=\ttfamily\footnotesize,
  backgroundcolor=\color{PromptGray},
  frame=single,
  rulecolor=\color{Muted},
  framerule=0.3pt,
  breaklines=true,
  breakatwhitespace=false,
  columns=fullflexible,
  keepspaces=true,
  showstringspaces=false,
  aboveskip=2pt,
  belowskip=5pt
}

\begin{document}
\nolinenumbers

\title{Can Compact Open-Weight Large Language Models Recover Mobility Events from Incomplete Trajectories? A Controlled Evaluation Using Air Trips}

\TRBauthor{Hsuan-Chih Wang}{Graduate Research Assistant\\Department of Civil and Environmental Engineering, University of Maryland}{hcwang@umd.edu}[College Park, MD, 20742][0000-0002-8669-7915]
\AuthorHeaders{Wang and Yang}

\TRBauthor{Xianfeng Terry Yang, Ph.D.}{Associate Professor \& Director \\Maryland Transportation Institute \\Department of Civil and Environmental Engineering, University of Maryland}{xtyang@umd.edu}[College Park, MD, 20742][0000-0002-9416-6882]
\AuthorHeaders{Wang and Yang}

\maketitle
\noindent\textit{Accepted for presentation at the 106th Annual Meeting of the Transportation Research Board, January 2027.}

\section{Abstract}
\noindent\textbf{Objectives:}
Incomplete mobility trajectories can obscure the observations needed to identify important travel events and their endpoints. This study evaluates whether compact, locally deployed large language models can recover missing mobility information from incomplete trajectories, using high-confidence air trips as a controlled test case. It also examines whether more explicit prompts can compensate for the loss of endpoint evidence.

\hfill\break%
\noindent\textbf{Methods:}
The study used nationwide location-based services (LBS) data collected in the United States in January 2022 and airport reference data containing airport codes and coordinates. Thirty high-confidence air trips were converted into three incomplete-trajectory conditions by removing observations near the origin airport, the destination airport, or both. Four prompt variants and two large language models generated 720 requests, each using a two-stage procedure to identify the flight transition segment and then selected the origin and destination airports. Performance metrics include recovery rates for transitions, origins, destinations, exact airport pairs, and joint transition–airport-pair identification.

\hfill\break%
\noindent\textbf{Findings:}
Exact transition accuracy is 9.2\%, while origin and destination accuracy are 31.4\% and 10.7\%, respectively. Only 1.9\% of requests recover both the rule-identified transition and the reference airport pair. The most explicit prompt produced the highest observed exact transition and exact airport-pair recovery rates.

\hfill\break%
\noindent\textbf{Novelty:}
The study introduces a controlled endpoint-removal framework for evaluating recovery of discrete mobility events from incomplete trajectories. Air trips provide a measurable test case with identifiable transitions and endpoints, allowing failures to be examined by inference stage.

\hfill\break%
\noindent\textbf{Practical Applications:}
The framework can support the evaluation of trajectory inference methods before deployment. By distinguishing transition identification from endpoint assignment, it helps analysts determine whether recovered information should be retained, independently verified, or left unresolved.

\newpage
\section{Introduction}
Passively collected location-based services (LBS) data provide large-scale observations of human movement and support transportation applications such as trip extraction, travel-mode inference, origin--destination analysis, and estimation for travel demand. However, unlike continuously recorded trajectories, LBS data consist of irregular observations generated by mobile applications. Their frequency can vary across devices, applications, vendors, locations, and times. Important movements may therefore occur during periods in which no location is recorded, leaving the temporal boundaries and geographic endpoints of a mobility event uncertain. Recovering such event information from incomplete trajectories is important because errors introduced during reconstruction may propagate into subsequent estimates of travel behavior and transportation demand.

Previous studies have demonstrated that passive location data can support passenger aviation analysis and large-scale identification of trip ends and transportation modes \cite{Marzuoli2019,Yang2022}. Nevertheless, missing observations can obscure trip boundaries, hide intermediate movements, and remove the observations that are most informative for identifying a transportation mode. Air trips provide a useful controlled case for examining this problem because they represent discrete mobility events with identifiable transitions, travel direction, and geographic endpoints. A long and rapid movement may be consistent with air travel, but exact recovery requires identifying the observation interval associated with the flight and assigning the correct origin and destination airports. When observations near one or both airports are missing, these event attributes must be inferred from the remaining trajectory context.

Large language models (LLMs) provide a possible approach for combining the temporal, kinematic, and geographic evidence contained in mobility trajectories. Their ability to interpret structured context and integrate multiple forms of evidence makes them potentially useful for trajectory inference. Recent research has also evaluated LLMs for reconstructing masked trajectory segments from geographic and network context \cite{Truong2025}. However, a fluent or geographically plausible response does not necessarily represent exact recovery of the underlying mobility event. Reliability must be evaluated against known event transitions and endpoints. High-confidence air trips provide an appropriate test case because their flight transitions and airport endpoints can be identified before selected observations are deliberately removed.

A further gap concerns how missing evidence is evaluated. Naturally incomplete trajectories do not reveal the observations that are absent, making it difficult to determine whether a failed inference is caused by information loss, transition identification, or airport assignment. Controlled endpoint removal provides a direct way to isolate this issue: begin with a high-confidence air trip, remove specified airport observations, and compare the inferred result with the original event. A controlled experiment is therefore needed to determine whether prompt modifications improve recovery under endpoint information loss.

The analysis aims to address the following research questions:
\begin{enumerate}
    \item Under controlled removal of endpoint observations, how accurately can compact open-weight LLMs recover the transition and endpoints of a discrete mobility event?

    \item How does observed recovery performance vary across endpoint-removal conditions and prompt variants?
\end{enumerate}

This study presents a controlled endpoint-removal stress test for compact, locally deployed open-weight LLMs. The experiment begins with high-confidence air trips for which the flight transition and origin--destination (OD) airports are observable. Sightings near the origin, destination, or both endpoints are then removed. Each masked trajectory is evaluated through a two-stage task workflow: Stage~1 selects the consecutive observations most likely to represent the flight transition, and Stage~2 assigns origin and destination airports to the selected endpoints. 

This study makes three main contributions. First, it introduces a controlled endpoint-removal framework for evaluating the recovery of discrete mobility events from incomplete trajectories, demonstrated using high-confidence air trips. Second, it decomposes recovery into transition identification, origin assignment, destination assignment, and complete event recovery so that error accumulation can be examined by inference stage. Third, it characterizes the reliability limits of compact open-weight LLMs and assesses whether alternative prompting strategies yield consistent improvements when endpoint evidence is missing.

\section{Literature Review}
Passive location data have supported trip extraction, travel-mode identification, and large-scale mobility estimation, but the underlying records generally contain only a device identifier, timestamp, and coordinates. Transportation attributes must therefore be inferred from movement patterns. Earlier GPS-based studies used change-point segmentation and engineered kinematic features to distinguish travel modes, while subsequent deep-learning approaches learned representations from sequences of speed, acceleration, jerk, and bearing information \cite{Zheng2010,Dabiri2018}. At a larger scale, mobile-device location data have been used to estimate multimodal travel shares, and mobile-phone records have supported passenger-centered analysis of the air transportation system \cite{Yang2022,Marzuoli2019}. These applications establish the value of passive mobility data but also show that their outputs depend on how trips, transitions, and endpoints are extracted.

Missing observations are a central limitation of passive trajectories. Sparse probe updates have motivated methods that reconstruct the most likely path between recorded points \cite{Wan2016}; sparse call-detail records can omit entire intermediate visits \cite{Zhao2022}; and recent imputation research explicitly models mobile-data missingness and its uncertainty \cite{Ugurel2024}. These studies primarily recover paths, coordinates, or visits. Air-trip recovery poses a related but distinct problem because the evaluation target is a discrete reference event: the correct observation pair must bracket the flight, and the two endpoints must be assigned to the correct airports. A trajectory that is merely plausible may still be incorrect under this stricter criterion.

LLM-based mobility research has begun to test whether general-purpose language models can interpret structured spatiotemporal histories. For example, LLM-Mob uses historical and recent stays for next-location prediction, while TSI-LLM infers activity sequences and other trajectory semantics from formatted mobility records \cite{Wang2023,Luo2024}. GeoLLM demonstrates that external geographic context can improve geospatial prediction, and GLOBALTRACE evaluates masked GPS-segment reconstruction using road-network context \cite{Manvi2024,Truong2025}. Collectively, this work suggests that LLMs can combine heterogeneous mobility and geographic evidence, but it does not establish that compact models can exactly recover a known flight transition and airport pair after the most diagnostic endpoint observations have been deleted.

Recent research provides one possible strategy for reducing task complexity. Chain-of-thought and least-to-most prompting decompose difficult problems into intermediate reasoning steps \cite{Wei2022,Zhou2023}. Their effectiveness, however, depends on model capacity, task formulation, and the evidence available in the prompt. By starting with high-confidence positive air trips, systematically deleting origin- and destination-proximate sightings, and scoring each inference component separately, the study aims to diagnose LLMs' reliability under controlled information loss.

\section{Method}

\begin{figure}[!htbp]
\centering
\includegraphics[width=\linewidth]{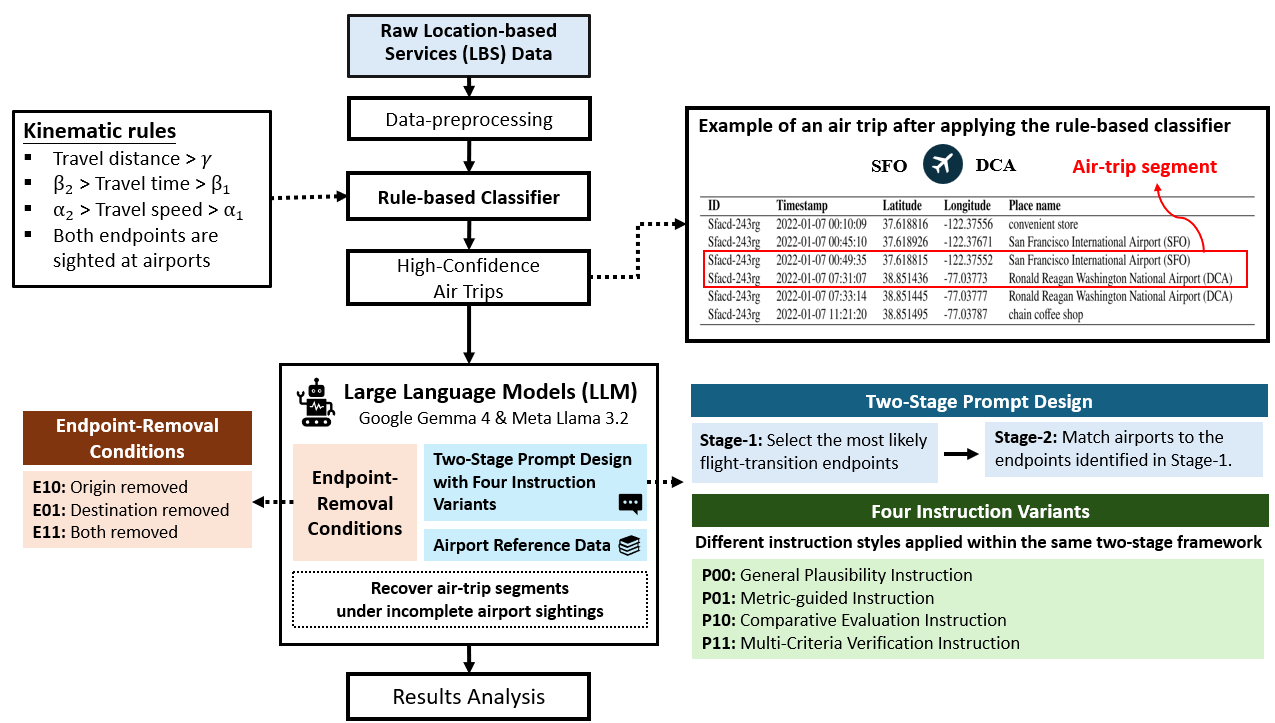}
\caption{Framework of the study}
\label{fig:pipeline}
\end{figure}

Figure~\ref{fig:pipeline} presents the study framework. The process begins with raw location-based services (LBS) data, which are chronologically organized and preprocessed to calculate the distance, elapsed time, and implied speed between consecutive observations. A rule-based classifier then applies the specified kinematic thresholds and verifies airport sightings at both endpoints to identify high-confidence air trips. These high-confidence air trips provide the reference air-trip segments used to construct the experimental inputs.

For each air trip, sightings near the origin, destination, or both endpoints are removed to create the three endpoint-removal conditions. Each resulting trajectory is evaluated under four prompt variants and supplied to an LLM together with the airport reference data. The two-stage prompting procedure first identifies the most likely flight-transition interval and then infers origin and destination airports to its endpoints. Model outputs are compared with the rule-identified transition and airport pair, and the results are analyzed across endpoint-removal conditions, prompt variants, and LLMs.

\subsection{Data Processing and Analytics}
Two datasets are used in this study: LBS data and airport reference data. 

\subsubsection{LBS Data}
This study uses a nationwide LBS dataset comprising observations collected from multiple data vendors in January 2022. Each observation includes a randomly hashed device identifier, latitude and longitude coordinates, and a timestamp. Table~\ref{tab:sample-lbs} presents a sample of these data. To protect privacy, noise has been added to all entries in the table; the sample is provided solely to illustrate the types of information contained in the LBS dataset. Before the rule-based air-trip classifier is applied and the trajectory context is provided to the LLMs, the distance, elapsed time, and implied speed between each pair of consecutive sightings are calculated and appended to the dataset.

\begin{table}[!t]
 \centering
 \caption{Sample of LBS Data}
 \label{tab:sample-lbs}
 \fontsize{10pt}{12pt}\selectfont
 \setlength{\tabcolsep}{3pt}
 \begin{tabularx}{\linewidth}{llllX}
 \hline
 \textbf{ID} & \textbf{Timestamp} & \textbf{Latitude} & \textbf{Longitude} & \textbf{Place name} \\
 \hline
 Sfacd-243rg & 2022-01-07 00:10:09 & 37.618816 & -122.37556 & convenience store \\
 Sfacd-243rg & 2022-01-07 00:45:10 & 37.618926 & -122.37671 & San Francisco International Airport (SFO) \\
 Sfacd-243rg & 2022-01-07 00:49:35 & 37.618815 & -122.37552 & San Francisco International Airport (SFO) \\
 Sfacd-243rg & 2022-01-07 07:31:07 & 38.851436 & -77.03773 & Ronald Reagan Washington National Airport (DCA) \\
 Sfacd-243rg & 2022-01-07 07:33:14 & 38.851445 & -77.03777 & Ronald Reagan Washington National Airport (DCA) \\
  Sfacd-243rg & 2022-01-07 11:21:20 & 38.851495 & -77.03787 & chain coffee shop \\
 \hline
 \end{tabularx}
\end{table}

\subsubsection{Airport Reference Data}
Because the LLMs are deployed locally without external data access, an airport reference dataset is provided as part of the input for air-trip recovery. The dataset contains three fields for each airport: the International Air Transport Association (IATA) code, latitude, and longitude. For example, the record \texttt{ABE, 40.652411, -75.435767} identifies Lehigh Valley International Airport by its IATA code and geographic coordinates. The IATA code uniquely identifies each airport, while the coordinates allow the model to assess the spatial consistency between observed device locations and candidate airports.

\subsection{Rule-based Classifier for Air Trips}\label{subsec:preprocess}
Potential air trips are identified using a rule-based classifier with thresholds for travel duration, distance, and speed:

\begin{itemize}
  \item \textbf{Travel duration:} The elapsed travel time must be between 30 minutes ($\beta_{1}$) and 18 hours ($\beta_{2}$). The 30-minute lower bound follows the air-trip classification threshold specified in the Next Generation NHTS Passenger Origin-Destination Data \cite{NextGenNHTSOD}. The 18-hour upper bound accommodates long domestic U.S. air trips as well as time associated with airport access, waiting, and egress.

  \item \textbf{Travel distance:} The observed displacement must exceed 50 miles ($\gamma$), following the air-trip classification threshold specified in the Next Generation NHTS Passenger Origin-Destination Data \cite{NextGenNHTSOD}.

  \item \textbf{Travel speed:} The estimated travel speed must be between 100 mph ($\alpha_{1}$) and 800 mph ($\alpha_{2}$). The lower bound follows the Next Generation NHTS air-trip classification threshold, whereas the upper bound represents the upper range of commercial-aircraft speeds \cite{NextGenNHTSOD,FAA2007}.

  \item \textbf{Airport endpoints:} Both endpoints must be sighted at two distinct airports.

\end{itemize}

\subsection{Endpoint-Removal Conditions}
To test recovery after loss of endpoint evidence, this study proposes three endpoint-removal conditions. Table~\ref{tab:conditions} describes the details of these three conditions. Each endpoint-removal version is generated independently from the original sighting data rather than sequentially from another version.

\begin{table}[!htbp]
\caption{Endpoint-Removal Conditions}\label{tab:conditions}
\centering
\begin{tabularx}{\linewidth}{llY}
\toprule
Code & Condition & Construction and intended evidence loss \\
\midrule
E10 & Origin removed & Removes observations within 20 mi of the origin endpoint during the 30 min preceding departure. \\
E01 & Destination removed & Removes observations within 20 mi of the destination endpoint during the 30 min following arrival. \\
E11 & Both removed & Removes the union of the origin and destination deletion sets. \\
\bottomrule
\end{tabularx}
\end{table}

The three masking conditions can be mathematically expressed as follows: Let $(t_i^O,s_i^O)$ and $(t_i^D,s_i^D)$ denote the origin (departure) and destination (arrival) endpoint times and locations, respectively, and let $d(\cdot,\cdot)$ be Haversine distance. The origin-removal set is

\begin{equation}
 \mathcal{R}_i^O=\left\{j:\;0\le t_i^O-t_{ij}\le30\minutes,\; d(s_{ij},s_i^O)\le20\mi\right\},
 \label{eq:originremove}
\end{equation}

and the destination-removal set is

\begin{equation}
 \mathcal{R}_i^D=\left\{j:\;0\le t_{ij}-t_i^D\le30\minutes,\; d(s_{ij},s_i^D)\le20\mi\right\}.
 \label{eq:destremove}
\end{equation}

The three endpoint-removal conditions are expressed as follows: 

\begin{align}
 \mathcal{T}_{i,E10}&=\mathcal{T}_i\setminus\mathcal{R}_i^O,\\
 \mathcal{T}_{i,E01}&=\mathcal{T}_i\setminus\mathcal{R}_i^D,\\
 \mathcal{T}_{i,E11}&=\mathcal{T}_i\setminus(\mathcal{R}_i^O\cup\mathcal{R}_i^D).
 \label{eq:conditions}
\end{align}

\begin{figure}[!ht]
\centering
\includegraphics[width=\linewidth]{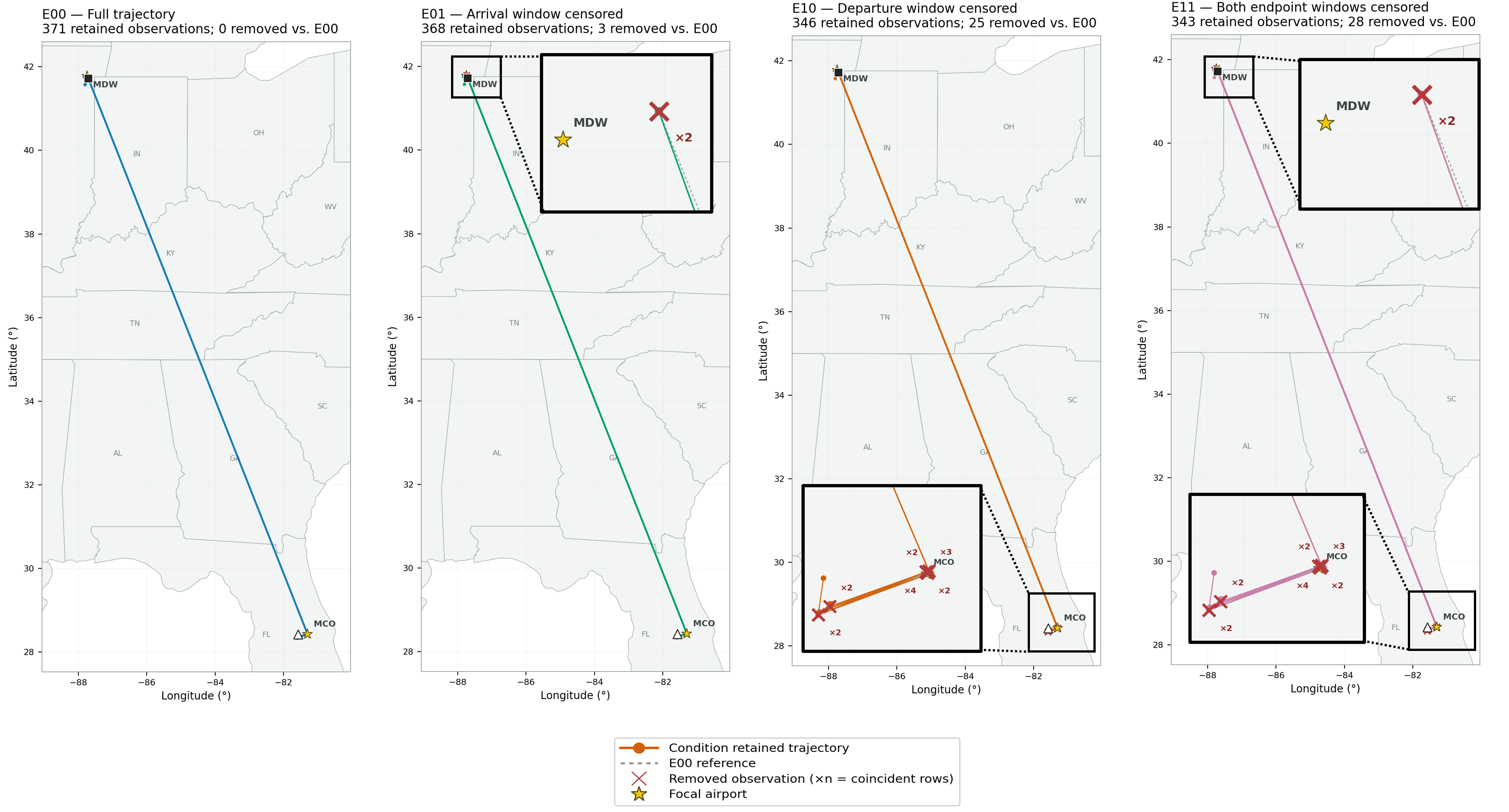}
\caption{Example of an air-trip trajectory before and after endpoint removal}
\label{fig:trj}
\end{figure}

Figure~\ref{fig:trj} illustrates an air-trip trajectory from Orlando International Airport (MCO) to Chicago Midway International Airport (MDW) before and after endpoint removal. The leftmost panel shows the complete trajectory. The remaining panels show the three endpoint-removal conditions: E01 removes three observations near the destination airport (MDW), E10 removes twenty-five observations near the origin airport (MCO), and E11 removes all twenty-eight observations across both endpoint windows. Notably, under E10, some observations remain in the vicinity of MCO after endpoint removal, but they do not provide direct evidence of a sighting at the airport. Together, the panels illustrate how endpoint removal reduces the airport-proximate evidence available for directly identifying the air trip's origin and destination.

\FloatBarrier

\subsection{Two-Stage Prompt Structures}
To reduce reasoning complexity and evaluate LLM recovery across experimental conditions, this study uses a two-stage prompting procedure that separates flight-transition selection from airport matching. Stage~1 asks each LLM to select the pair of consecutive observations that most likely represents the flight transition. After Stage~1, an external Python program creates a shortlist of the five airports nearest to each selected endpoint from the reference airport dataset. The procedure then proceeds to Stage~2, in which the LLM assigns origin and destination airports to the corresponding endpoints of the observation pair selected in Stage~1.

Four prompt variants---P00, P01, P10, and P11---are designed to move from a general plausibility assessment toward more explicit reasoning instructions. P00 uses general plausibility instructions. P01 emphasizes transition distance, speed, elapsed time, and endpoint-to-airport distance. P10 emphasizes comparisons among transition candidates. P11 combines broader transition checks with geographic verification during airport selection. Table~\ref{tab:prompts} summarizes the instructions applied at each stage of the four prompt variants.

\begin{figure}[p]
\centering
\includegraphics[width=0.80\linewidth]{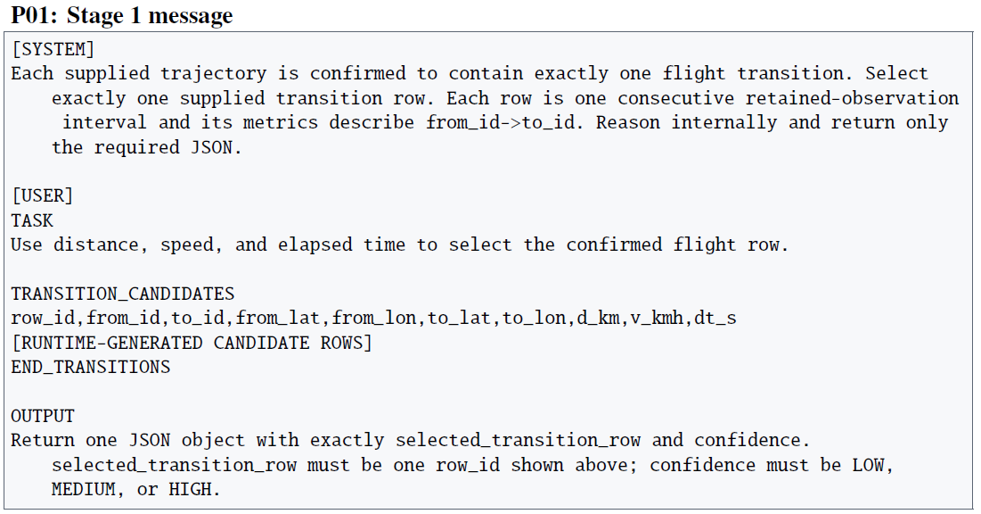}
\caption{P01 Stage-1 Full Prompt Text}
\label{fig:P01_Stage1}
\vspace{0.12in}
\includegraphics[width=0.80\linewidth]{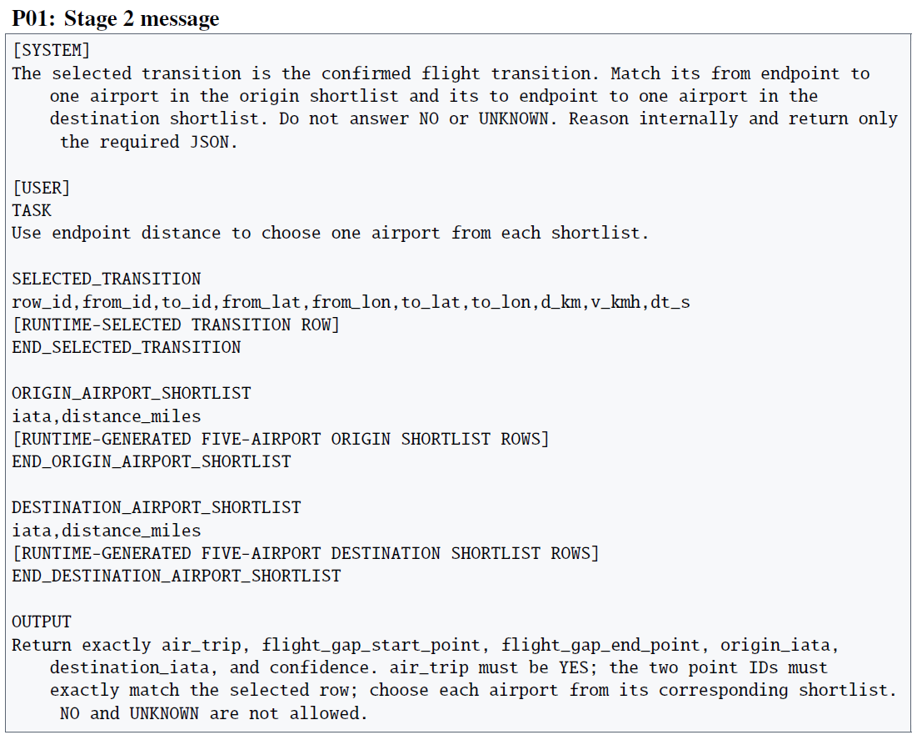}
\caption{P01 Stage-2 Full Prompt Text}
\label{fig:P01_Stage2}
\end{figure}

Examples of the complete prompts are presented in Figures~\ref{fig:P01_Stage1} and~\ref{fig:P01_Stage2}. The prompts require structured JSON responses and require the model to select from the supplied candidates. In particular, Stage~2 does not allow a \texttt{NO} or \texttt{UNKNOWN} response. Consequently, the experiments evaluate exact transition and airport assignment for referenced air trips rather than the model's ability to abstain from making an unreliable prediction.

\begin{table}[!htbp]
\caption{Summary of the Two-Stage Prompt Instructions}\label{tab:prompts}
\centering
\begin{tabularx}{\linewidth}{lYY}
\toprule
Code & Stage 1 instruction & Stage 2 instruction \\
\midrule
P00 & "Select the transition row that most plausibly represents an air trip." & "Select the nearest plausible origin and destination airports." \\
P01 & "Evaluate the distance, speed, and elapsed time when selecting the transition rows." & "Select airports using endpoint-to-airport distance." \\
P10 & "Compare all candidate rows before selecting the flight transition." & "Map the FROM endpoint to the origin and the TO endpoint to the destination." \\
P11 & "Select transition rows after checking timing,
distance, speed, and non-air alternatives." & "Select airports after verifying geographic proximity."\\
\bottomrule
\end{tabularx}
\end{table}

\section{Experimental Design}\label{sec:evaluation}

This study analyzes 30 high-confidence air-trip trajectories across the three endpoint-removal conditions in Table~\ref{tab:conditions} and the four prompt variants in Table~\ref{tab:prompts}. Two locally deployed open-weight LLMs are evaluated: Google \texttt{gemma4:e2b} \cite{googledeepmind2026gemma4} and Meta \texttt{llama3.2:3b-instruct-q4\_K\_M} \cite{meta2024llama32}. The resulting $30\times3\times4\times2$ design contains 720 logical requests. Each logical request uses two model calls, resulting in 1,440 stage calls. In both stages, each LLM uses a context window of approximately 32,000 tokens, the maximum output is 512 tokens, \texttt{temperature=1.0}, \texttt{top\_p}=0.95, \texttt{top\_k}=64, and a fixed random seed. Note that the 720 requests are repeated evaluations of the same 30 reference trips rather than 720 independent trajectories. Each distinct trip is therefore the underlying evaluation unit, while the request-level counts describe model responses across conditions, prompts, and LLMs.

Five measures are reported:
\begin{itemize}
  \item \textbf{Exact transition:} Stage~1 selects the same transition row as the reference trip. Airport correctness is not required.
  \item \textbf{Origin:} The inferred origin airport matches the reference origin.
  \item \textbf{Destination:} The inferred destination airport matches the reference destination.
  \item \textbf{Exact OD:} Both inferred airports match the reference pair, regardless of Stage~1 correctness.
  \item \textbf{Transition + OD:} The transition row, origin airport, and destination airport are all correct in the same request.
\end{itemize}

\section{Results}
\subsection{Overall and Endpoint-Removal Performance}
Table~\ref{tab:performance_condition} shows the results under the three endpoint-removal conditions. Exact transition recovery is 66/720 (9.2\%). Origin accuracy is 226/720 (31.4\%), whereas destination accuracy is 77/720 (10.7\%). Fourteen responses recover both airports, yielding an exact-OD rate of 14/720 (1.9\%). Every exact-OD recovery also selects the rule-identified transition, so joint transition and OD recovery is likewise 14/720 (1.9\%). Among all conditions, E10 has the highest observed destination and exact-OD rates. E11 has the highest transition and origin rates, while E01 has the lowest destination and exact-OD rates.

\begin{table}[!ht]
\caption{Performance by Endpoint-Removal Condition}\label{tab:performance_condition}
\centering
\begin{threeparttable}
\fontsize{11pt}{13pt}\selectfont
\begin{tabularx}{\linewidth}{lCCCCC}
\toprule
Condition & Exact transition & Origin & Destination & Exact OD & Transition + OD \\
\midrule
E10 & 23/240 (9.6\%) & 69/240 (28.8\%) & 33/240 (13.8\%) & 8/240 (3.3\%) & 8/240 (3.3\%) \\
E01 & 19/240 (7.9\%) & 73/240 (30.4\%) & 18/240 (7.5\%) & 1/240 (0.4\%) & 1/240 (0.4\%) \\
E11 & 24/240 (10.0\%) & 84/240 (35.0\%) & 26/240 (10.8\%) & 5/240 (2.1\%) & 5/240 (2.1\%) \\
\midrule
Overall & 66/720 (9.2\%) & 226/720 (31.4\%) & 77/720 (10.7\%) & 14/720 (1.9\%) & 14/720 (1.9\%) \\
\bottomrule
\end{tabularx}
\end{threeparttable}
\end{table}

The results do not combine into reliable end-to-end inference. Although the origin is selected correctly in nearly one-third of requests, the much lower destination and transition rates reduce exact-OD recovery to 1.9\%. The condition-level pattern is also not monotonic. For example, E11, which removes both origin and destination endpoints, has the highest observed transition and origin rates, whereas E10 has the highest destination and exact-OD rates.

\subsection{Performance Across Prompt Variants}
Table~\ref{tab:performance_prompt} shows results aggregated across endpoint-removal conditions and models. P11 has the highest observed rate for all five outcomes, including exact-transition recovery at 20/180 (11.1\%) and exact-OD recovery at 6/180 (3.3\%). P01 and P10 both recover 17/180 exact transitions (9.4\%), although P01 has slightly higher component-level airport accuracy. However, when it comes to exact-OD recovery, the performance remain limited. Even the most explicit prompt, P11, which has the highest observed exact-OD rate, recovers the correct airport pair in only 3.3\% of requests. These results indicate that more detailed instructions do not substantially overcome missing endpoint evidence in most cases.

\begin{table}[!htbp]
\caption{Performance Across Prompt Variants}\label{tab:performance_prompt}
\centering
\fontsize{10pt}{12pt}\selectfont
\begin{tabularx}{\linewidth}{lCCCCC}
\toprule
Prompt & Exact transition & Origin & Destination & Exact OD & Transition + OD \\
\midrule
P00 & 12/180 (6.7\%) & 53/180 (29.4\%) & 16/180 (8.9\%) & 3/180 (1.7\%) & 3/180 (1.7\%) \\
P01 & 17/180 (9.4\%) & 58/180 (32.2\%) & 20/180 (11.1\%) & 2/180 (1.1\%) & 2/180 (1.1\%) \\
P10 & 17/180 (9.4\%) & 54/180 (30.0\%) & 19/180 (10.6\%) & 3/180 (1.7\%) & 3/180 (1.7\%) \\
P11 & 20/180 (11.1\%) & 61/180 (33.9\%) & 22/180 (12.2\%) & 6/180 (3.3\%) & 6/180 (3.3\%) \\
\midrule
Overall & 66/720 (9.2\%) & 226/720 (31.4\%) & 77/720 (10.7\%) & 14/720 (1.9\%) & 14/720 (1.9\%) \\
\bottomrule
\end{tabularx}
\end{table}

\subsection{Performance by Endpoint-Removal Condition and Prompt Variant}
Table~\ref{tab:condition_prompt_od} reports exact-OD recovery for each condition--prompt combination. E10--P11 has the highest observed rate, with 4/60 recoveries (6.7\%), whereas E01 has no exact-OD recovery under P00, P01, or P10. Because all 14 exact-OD recoveries also select the rule-identified transition, the joint transition-plus-OD results are numerically identical and are not reported separately. The advantage of E10--P11 is based on only four successful recoveries, while most condition--prompt results contain no more than two. These sparse outcomes indicate generally low recovery performance across prompt designs under endpoint removal.

\begin{table}[!htbp]
\caption{Exact-OD Recovery by Endpoint-Removal Condition and Prompt}\label{tab:condition_prompt_od}
\centering
\small
\begin{tabularx}{0.82\linewidth}{lCCCC}
\toprule
Condition & P00 & P01 & P10 & P11 \\
\midrule
E10 & 2/60 (3.3\%) & 1/60 (1.7\%) & 1/60 (1.7\%) & 4/60 (6.7\%) \\
E01 & 0/60 (0.0\%) & 0/60 (0.0\%) & 0/60 (0.0\%) & 1/60 (1.7\%) \\
E11 & 1/60 (1.7\%) & 1/60 (1.7\%) & 2/60 (3.3\%) & 1/60 (1.7\%) \\
\bottomrule
\end{tabularx}
\end{table}
\FloatBarrier

When examined separately by model, Gemma recovered both airports in 10/360 (2.8\%), compared with 4/360 (1.1\%) for Llama. The largest observed separation occurred under E10--P11, where Gemma recovered 4/30 (13.3\%) and Llama recovered 0/30.

\begin{table}[!ht]
\centering
\caption{Exact-OD recovery by model, endpoint-removal condition, and prompt}
\small
\setlength{\tabcolsep}{5.5pt}
\renewcommand{\arraystretch}{1.22}
\begin{tabular}{llcccc}
\hline
\textbf{Condition} & \textbf{Model} & \textbf{P00} & \textbf{P01} & \textbf{P10} & \textbf{P11} \\
\hline
E10 & Gemma & 1/30 (3.3\%) & 1/30 (3.3\%) & 1/30 (3.3\%) & 4/30 (13.3\%) \\
E10 & Llama & 1/30 (3.3\%) & 0/30 (0.0\%) & 0/30 (0.0\%) & 0/30 (0.0\%) \\
\hline
E01 & Gemma & 0/30 (0.0\%) & 0/30 (0.0\%) & 0/30 (0.0\%) & 1/30 (3.3\%) \\
E01 & Llama & 0/30 (0.0\%) & 0/30 (0.0\%) & 0/30 (0.0\%) & 0/30 (0.0\%) \\
\hline
E11 & Gemma & 1/30 (3.3\%) & 0/30 (0.0\%) & 1/30 (3.3\%) & 0/30 (0.0\%) \\
E11 & Llama & 0/30 (0.0\%) & 1/30 (3.3\%) & 1/30 (3.3\%) & 1/30 (3.3\%) \\
\hline
\end{tabular}
\par\smallskip
\end{table}

\subsection{Stage 1 and Stage 2 Outcomes by Endpoint-Removal Condition}

Table~\ref{tab:stage-propagation} provides an insight into the stages at which recovery generally failed. Only 66/720 (9.2\%) responses passed Stage~1 by selecting the exact transition. The Stage~1 pass rate is 9.6\% (23/240) in E10, 7.9\% (19/240) in E01, and 10.0\% (24/240) in E11. Conditional on passing Stage~1, 52/66 (78.8\%) responses fail at Stage~2, leaving 14/66 (21.2\%) with exact OD recovery. The conditional Stage~2 failure rate is highest in E01 at 18/19 (94.7\%), followed by E11 at 19/24 (79.2\%) and E10 at 15/23 (65.2\%). These results indicate that most recovery loss occurred at Stage~1, while Stage~2 introduces further losses among responses that have selected the correct transition.

\begin{table}[htbp]
\centering
\fontsize{10pt}{12pt}\selectfont
\caption{Stage 1 and Stage 2 outcomes by endpoint-removal condition}
\label{tab:stage-propagation}
\setlength{\tabcolsep}{10pt}
\renewcommand{\arraystretch}{1.35}
\begin{tabularx}{\linewidth}{lCCCC}
\toprule
Condition &
S1 failure&
S1 pass &
\shortstack{S2 failure $\mid$\\S1 pass} &
\shortstack{S2 pass $\mid$\\S1 pass} \\
\midrule
E10     & 217/240 (90.4\%) & 23/240 (9.6\%)  & 15/23 (65.2\%) & 8/23 (34.8\%) \\
E01     & 221/240 (92.1\%) & 19/240 (7.9\%)  & 18/19 (94.7\%) & 1/19 (5.3\%) \\
E11     & 216/240 (90.0\%) & 24/240 (10.0\%) & 19/24 (79.2\%) & 5/24 (20.8\%) \\
Overall & 654/720 (90.8\%) & 66/720 (9.2\%)  & 52/66 (78.8\%) & 14/66 (21.2\%) \\
\bottomrule
\end{tabularx}

\begin{minipage}{0.95\linewidth}
\fontsize{10pt}{12pt}\selectfont
\vspace{4pt}
\textit{Note.} "S1 pass" denotes selection of the exact transition.
"S2 pass" denotes exact recovery of both airports in the correct
origin--destination order. The Stage~2 columns are conditional on
"S1 pass" and therefore sum to 100\% within each row.
\end{minipage}
\end{table}

\section{Discussion and Conclusion}
This study proposes a framework for evaluating whether locally deployed open-weight LLMs can recover air-trip transitions and airport pairs from systematically masked LBS trajectories. The framework first identifies high-confidence air trips using kinematic and airport-endpoint rules and then removes observations near the origin, destination, or both endpoints. Stage~1 selects the pair of consecutive observations that most likely represents the flight transition, while Stage~2 assigns origin and destination airports to the corresponding endpoints. This two-stage design reports transition and airport outcomes separately and reveals how errors can propagate across the inference pipeline. Because the airport shortlists are generated from the endpoints selected in Stage~1, the airport measures reflect both upstream transition errors and downstream airport-selection errors.

The controlled endpoint-removal experiment reveals a substantial reliability gap in air-trip inference by compact LLMs. Across 720 requests, exact-transition accuracy is 9.2\%, origin accuracy is 31.4\%, and destination accuracy is 10.7\%. Only 14 responses recover the complete airport pair, resulting in exact-OD and joint transition-plus-OD rates of 1.9\%. The comparison of performance across Stage~1 and Stage~2 further identifies Stage~1 as the primary bottleneck, as approximately 90\% of requests failed to select the exact transition segment. Performance is also non-monotonic across endpoint-removal conditions: E11 produces the highest observed transition and origin rates, whereas E10 produces the highest destination and exact-OD rates. These patterns do not establish a consistent relationship between the location of removed evidence and recovery performance. Overall, prompt engineering alone does not compensate for missing endpoint information in this setting; compact LLM outputs should therefore be treated as hypotheses requiring external validation rather than as recovered air-trip records.

For transportation applications, these results argue against using a compact LLM as an unconditional imputation module for missing airport sightings. Model outputs would require external safeguards, including kinematic and flight-schedule checks, validation of airport candidate sets, calibrated confidence, and the ability to abstain when evidence is insufficient. The contribution of the proposed framework is therefore diagnostic rather than operational: controlled masking and stage-specific metrics expose reliability limits that may be hidden when models are evaluated only on complete trajectories or judged by whether their outputs appear plausible.

Several limitations remain. First, both evaluated models, Gemma and Llama, are open-weight LLMs deployed locally. Their compact sizes and local inference configurations may constrain their reasoning capabilities relative to larger models. Therefore, the findings may not generalize to larger LLMs, such as ChatGPT, Claude or Gemini. Second, the evaluation includes only 30 trips. A larger sample spanning more air routes and observation patterns is needed to produce more precise and generalizable performance estimates. Finally, evaluations for additional LLMs are needed before operational deployment can be considered.

\section{Authors' Note}
The authors acknowledge that the large language model ChatGPT 5.5 was used to support manuscript grammar editing and code troubleshooting during manuscript preparation. The authors recognize the limitations of language models and rigorously verify the accuracy, validity, and appropriateness of the resulting text, code, and analyses.

\section*{AUTHOR CONTRIBUTIONS}
The authors confirm contribution to the paper as follows: 
study conception and design: Hsuan-Chih Wang and Xianfeng Terry Yang, data analysis and interpretation of results: Hsuan-Chih Wang
draft manuscript preparation: Hsuan-Chih Wang and Xianfeng Terry Yang. All authors reviewed the results and approved the final version of the manuscript.

\section*{DECLARATION OF CONFLICTING INTERESTS}
The authors declare no potential conflicts of interest with respect to the research, authorship, and publication of this article. 

\section*{FUNDING}
The authors received no financial support for the research, authorship, and/or publication of this article.

\printbibliography[title={REFERENCES}]

@article{Marzuoli2019,
  author  = {Marzuoli, Aude and Boidot, Emmanuel and Feron, Eric and Srivastava, Ashok N.},
  title   = {Implementing and validating air passenger-centric metrics using mobile phone data},
  journaltitle = {Journal of Aerospace Information Systems},
  year    = {2019},
  volume  = {16},
  number  = {4},
  pages   = {132--147},
  doi     = {10.2514/1.I010625}
}

@article{Yang2022,
  author  = {Yang, Mofeng and Pan, Yixuan and Darzi, Aref and Ghader, Sepehr and Xiong, Chenfeng and Zhang, Lei},
  title   = {A data-driven travel mode share estimation framework based on mobile device location data},
  journaltitle = {Transportation},
  year    = {2022},
  volume  = {49},
  number  = {5},
  pages   = {1339--1383},
  doi     = {10.1007/s11116-021-10214-3}
}

@article{Zheng2010,
  author  = {Zheng, Yu and Chen, Yukun and Li, Quannan and Xie, Xing and Ma, Wei-Ying},
  title   = {Understanding transportation modes based on {GPS} data for web applications},
  journaltitle = {ACM Transactions on the Web},
  year    = {2010},
  volume  = {4},
  number  = {1},
  pages   = {1--36},
  doi     = {10.1145/1658373.1658374}
}

@article{Dabiri2018,
  author  = {Dabiri, Sina and Heaslip, Kevin},
  title   = {Inferring transportation modes from {GPS} trajectories using a convolutional neural network},
  journaltitle = {Transportation Research Part C: Emerging Technologies},
  year    = {2018},
  volume  = {86},
  pages   = {360--371},
  doi     = {10.1016/j.trc.2017.11.021}
}

@article{Wan2016,
  author  = {Wan, Nianfeng and Vahidi, Ardalan and Luckow, Andre},
  title   = {Reconstructing maximum likelihood trajectory of probe vehicles between sparse updates},
  journaltitle = {Transportation Research Part C: Emerging Technologies},
  year    = {2016},
  volume  = {65},
  pages   = {16--30},
  doi     = {10.1016/j.trc.2016.01.010}
}

@article{Zhao2022,
  author  = {Zhao, Zhan and Koutsopoulos, Haris N. and Zhao, Jinhua},
  title   = {Identifying hidden visits from sparse call detail record data},
  journaltitle = {Transactions in Urban Data, Science, and Technology},
  year    = {2022},
  volume  = {1},
  number  = {3--4},
  pages   = {121--141},
  doi     = {10.1177/27541231221124164}
}

@article{Ugurel2024,
  author  = {U{\u{g}}urel, Ekin and Guan, Xiangyang and Wang, Yanchao and Huang, Shuai and Wang, Qi and Chen, Cynthia},
  title   = {Correcting missingness in passively generated mobile data with multi-task {Gaussian} processes},
  journaltitle = {Transportation Research Part C: Emerging Technologies},
  year    = {2024},
  volume  = {161},
  eid     = {104523},
  doi     = {10.1016/j.trc.2024.104523}
}

@article{Wang2023,
  author        = {Wang, Xinglei and Fang, Meng and Zeng, Zichao and Cheng, Tao},
  title         = {Where would {I} go next? Large language models as human mobility predictors},
  journaltitle  = {arXiv preprint arXiv:2308.15197},
  year          = {2023},
  eprint        = {2308.15197},
  archivePrefix = {arXiv},
  primaryClass  = {cs.AI},
  doi           = {10.48550/arXiv.2308.15197}
}

@article{Luo2024,
  author        = {Luo, Yuxiao and Cao, Zhongcai and Jin, Xin and Liu, Kang and Yin, Ling},
  title         = {Deciphering human mobility: Inferring semantics of trajectories with large language models},
  journaltitle  = {arXiv preprint arXiv:2405.19850},
  year          = {2024},
  eprint        = {2405.19850},
  archivePrefix = {arXiv},
  primaryClass  = {cs.AI},
  doi           = {10.48550/arXiv.2405.19850}
}

@inproceedings{Manvi2024,
  author    = {Manvi, Rohin and Khanna, Samar and Mai, Gengchen and Burke, Marshall and Lobell, David B. and Ermon, Stefano},
  title     = {{GeoLLM}: Extracting geospatial knowledge from large language models},
  booktitle = {The Twelfth International Conference on Learning Representations},
  year      = {2024},
  url       = {https://proceedings.iclr.cc/paper/2024/hash/a87f2df7c4ab0213c6ea228e7b7f0a4d-Abstract-Conference.html}
}

@article{Truong2025,
  author        = {Truong, Thinh Hung and Lau, Jey Han and Qi, Jianzhong},
  title         = {Understanding the geospatial reasoning capabilities of {LLMs}: A trajectory recovery perspective},
  journaltitle  = {arXiv preprint arXiv:2510.01639},
  year          = {2025},
  eprint        = {2510.01639},
  archivePrefix = {arXiv},
  primaryClass  = {cs.AI},
  doi           = {10.48550/arXiv.2510.01639}
}

@inproceedings{Wei2022,
  author    = {Wei, Jason and Wang, Xuezhi and Schuurmans, Dale and Bosma, Maarten and Ichter, Brian and Xia, Fei and Chi, Ed H. and Le, Quoc V. and Zhou, Denny},
  title     = {Chain-of-thought prompting elicits reasoning in large language models},
  booktitle = {Advances in Neural Information Processing Systems},
  year      = {2022},
  volume    = {35},
  pages     = {24824--24837},
  url       = {https://proceedings.neurips.cc/paper_files/paper/2022/hash/9d5609613524ecf4f15af0f7b31abca4-Abstract.html}
}

@inproceedings{Zhou2023,
  author    = {Zhou, Denny and Sch{\"a}rli, Nathanael and Hou, Le and Wei, Jason and Scales, Nathan and Wang, Xuezhi and Schuurmans, Dale and Cui, Claire and Bousquet, Olivier and Le, Quoc V. and Chi, Ed H.},
  title     = {Least-to-most prompting enables complex reasoning in large language models},
  booktitle = {The Eleventh International Conference on Learning Representations},
  year      = {2023},
  url       = {https://openreview.net/forum?id=WZH7099tgfM}
}

@techreport{NextGenNHTSOD,
  author      = {Zhang, Lei and Darzi, Aref and Pan, Yixuan and Yang, Mofeng and Sun, Qianqian and Kabiri, Aliakbar and Zhao, Guangchen and Ashoori, Mohammad and Namadi, Saeed Saleh and Xiong, Chenfeng and Jordan, Gregory W. and Stewart, Kathleen and Pack, Michael L.},
  title       = {Next Generation National Household Travel Survey National Origin--Destination Data: Passenger Origin--Destination Data Methodology Documentation},
  institution = {Federal Highway Administration, U.S. Department of Transportation},
  location    = {Washington, DC},
  date        = {2021-10},
  note        = {Revised April 2023},
  url         = {https://nhts.ornl.gov/od/assets/doc/2020_NextGen_NHTS_Passenger_OD_Data_Methodology_v2.pdf}
}

@manual{FAA2007,
  author       = {{Federal Aviation Administration}},
  title        = {Pilot's Handbook of Aeronautical Knowledge},
  organization = {U.S. Department of Transportation, Federal Aviation Administration},
  location     = {Washington, DC},
  year         = {2007},
  note         = {FAA-H-8083-25A}
}

@online{googledeepmind2026gemma4,
  author       = {{Google DeepMind}},
  title        = {Gemma 4 Model Card},
  organization = {Google AI for Developers},
  date         = {2026-07-16},
  url          = {https://ai.google.dev/gemma/docs/core/model_card_4},
  urldate      = {2026-07-30}
}

@online{meta2024llama32,
  author       = {{Meta AI}},
  title        = {Llama 3.2: Revolutionizing Edge {AI} and Vision with Open, Customizable Models},
  organization = {Meta AI},
  date         = {2024-09-25},
  url          = {https://ai.meta.com/blog/llama-3-2-connect-2024-vision-edge-mobile-devices/},
  urldate      = {2026-07-30}
}

\end{document}